%% file: main.tex
\documentclass[letterpaper, 10pt, conference]{ieeeconf}

\IEEEoverridecommandlockouts 

\usepackage{amsmath,amssymb,amsfonts}

\usepackage[pdftex]{graphicx}
\graphicspath{{/figures/}}
\DeclareGraphicsExtensions{.pdf,.jpeg,.png}

\renewcommand{\baselinestretch}{0.97} 

\usepackage{svg}

\usepackage{textcomp}
\usepackage{xcolor}
\usepackage[acronym]{glossaries}
\makeglossaries
\usepackage{algorithm}
\usepackage{algpseudocode}
\usepackage[hidelinks]{hyperref}

\def\BibTeX{{\rm B\kern-.05em{\sc i\kern-.025em b}\kern-.08em
    T\kern-.1667em\lower.7ex\hbox{E}\kern-.125emX}}

\usepackage{cite} 

\newacronym{rl}{RL}{Reinforcement Learning}
\newacronym{vqace}{VQ-ACE}{Vector Quantized Action Chunking Embedding}
\newacronym{vae}{VAE}{variational autoencoder}
\newacronym{cvae}{CVAE}{Conditional VAE}
\newacronym{mpc}{MPC}{model predictive control}
\newacronym{ema}{EMA}{exponential moving average}
\newacronym{dof}{DoF}{degrees of freedom}

\begin{document}


\title{\LARGE \bf 
VQ-ACE: Efficient Policy Search for Dexterous Robotic Manipulation via Action Chunking Embedding
}
%

\author{Chenyu Yang$^{1}$,
    Davide Liconti$^{1}$,
    Robert K. Katzschmann$^{1}$
\thanks{$^{1}$Soft Robotics Lab, ETH Zurich, Switzerland}%
    \thanks{{\tt\footnotesize \{\href{mailto:chenyang@ethz.ch}{chenyang}, \href{mailto:dliconti@ethz.ch}{dliconti}, \href{mailto:rkk@ethz.ch}{rkk}\}@ethz.ch}}}


\maketitle

\begin{abstract}
\input{00_abstract}
\end{abstract}



\input{01_introduction}

\input{02_related_works}

\input{03_methods}

\input{04_experiments}

\input{05_conclusion}


\section*{Acknowledgment}

The authors thank Yasunori Toshimitsu for his valuable contributions to reinforcement learning (RL) tasks, Adrian Hess for his work on the model predictive control (MPC) pipeline, and Sebastiano Oliani for his assistance with the cube stacking task. This work was supported by the SNSF Project Grant \#200021\_\,215489 and the Swiss Data Science Center (SDSC) Grant “FastPoints2Mesh” \#C22-08. Their support was instrumental in the development of this research.

\addtolength{\textheight}{-0cm}   

\bibliographystyle{IEEEtran}
\bibliography{IEEEabrv,reference}
%

\end{document}

%% file: 00_abstract.tex
Dexterous robotic manipulation remains a significant challenge due to the high dimensionality and complexity of hand movements required for tasks like in-hand manipulation and object grasping. This paper addresses this issue by introducing Vector Quantized Action Chunking Embedding (VQ-ACE), a novel framework that compresses human hand motion into a quantized latent space, significantly reducing the action space's dimensionality while preserving key motion characteristics. By integrating VQ-ACE with both Model Predictive Control (MPC) and Reinforcement Learning (RL), we enable more efficient exploration and policy learning in dexterous manipulation tasks using a biomimetic robotic hand.
Our results show that latent space sampling with MPC produces more human-like behavior in tasks such as Ball Rolling and Object Picking, leading to higher task success rates and reduced control costs. For RL, action chunking accelerates learning and improves exploration, demonstrated through faster convergence in tasks like cube stacking and in-hand cube reorientation. These findings suggest that VQ-ACE offers a scalable and effective solution for robotic manipulation tasks involving complex, high-dimensional state spaces, contributing to more natural and adaptable robotic systems. Code, data, and training details will be publicly available at    \href{https://srl-ethz.github.io/page-vq-ace/}{srl-ethz.github.io/page-vq-ace/}.

%% file: 01_introduction.tex
\section{Introduction}
Designing systems that mimic the dexterity and adaptability of the human hand is vital in bridging the gap between specialized robotic tools and more versatile systems capable of generalist intelligence.  
The human hand is a remarkable tool for interaction and expression, and plays a key role in intelligence. With its 27 \gls{dof}, the hand is capable of an immense variety of movements and postures, enabling fine motor skills that are critical for tasks ranging from delicate manipulation to powerful grasps. However, despite this anatomical complexity, humans do not control each degree of freedom independently with ease. Instead, the control of hand movements is often characterized by coupled motions, where multiple joints move in coordinated patterns. This empirical observation highlights the need to find a more appropriate low-dimensional representation of the motion of human hands.

We hypothesize that a good representation of hand motion should capture the following essential characteristics. This representation must be dynamic, reflecting the continuous nature of hand movements rather than static postures. It must also be compact, as human hand movements occupy only a small manifold within the vast space of possible trajectories. Despite this compactness, the representation must remain expressive enough to capture the full range of human-hand capabilities. Additionally, for practical implementation in autoregressive models, the representation should be quantized to facilitate efficient computation and learning. These considerations drive the need for an encoder-decoder network that can effectively compress hand motion sequences into a latent space, reducing the dimensionality of the action space while preserving the essential features required for robotic policy learning.

In this work we introduce \gls{vqace}, learn a compact, discrete latent space for human hand action sequences with a vector-quantized conditional auto-encoder. This latent space enables robotic systems to efficiently search for optimal policies with anthropomorphic priors.
We propose two applications utilizing this latent space: latent sampling ~\gls{mpc} and action chunked ~\gls{rl}. ). Both control algorithms benefit from efficient sampling and exploration during policy search, generating more natural, human-like behaviors in robotic hands.
Please refer to the supplementary material for the robot videos.

\begin{figure}[t]
    \centering
    \includegraphics[width=0.9\linewidth]{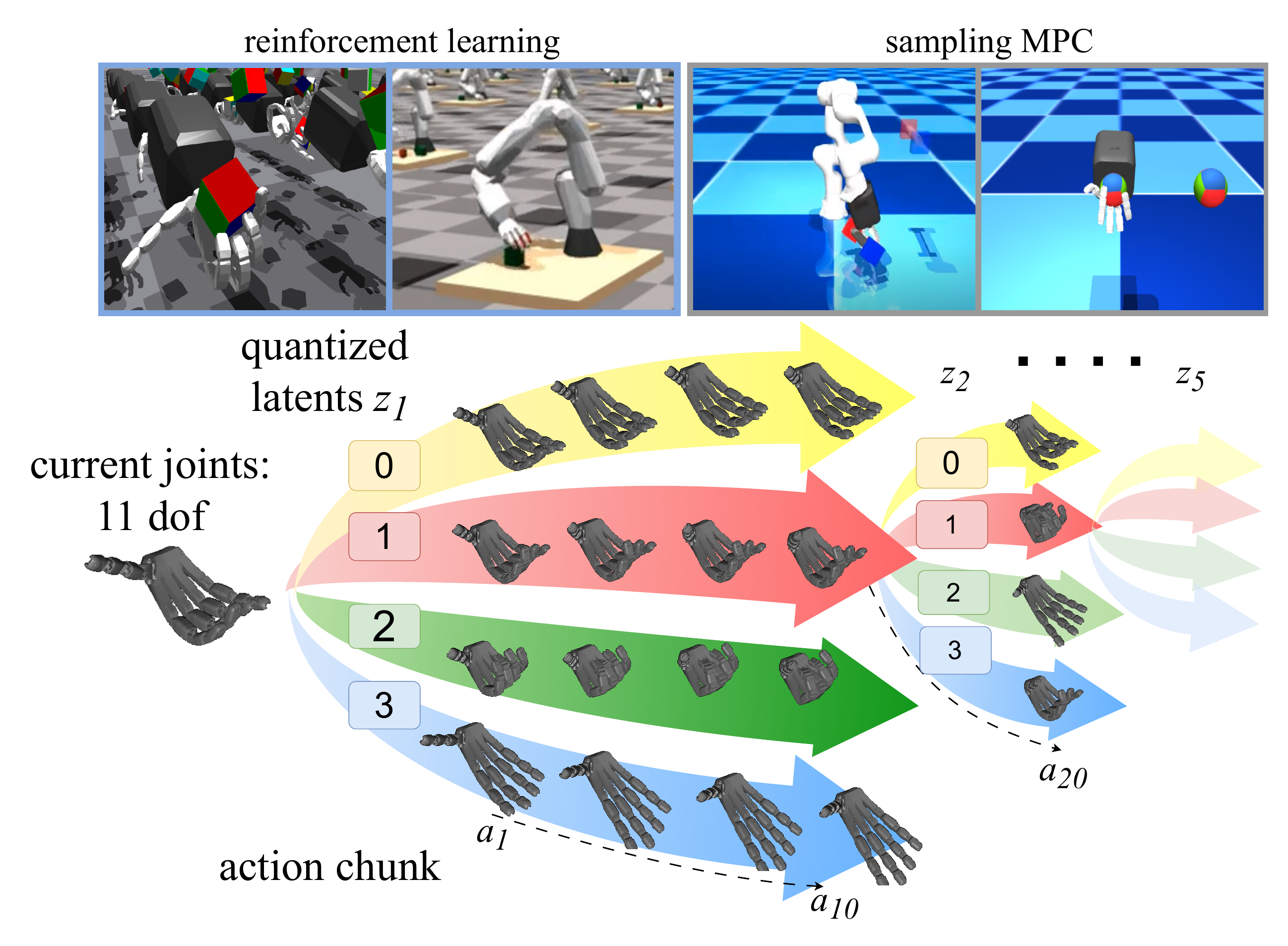}
    \caption{We introduce \gls{vqace}, a method that learns a compact representation of complex human hand motion in a lower-dimensional space. For a 1-second action chunk with 11 \gls{dof}, our approach encodes it into 5 tokens, each taking one of 4 possible discrete values. This learned latent space can be used in both sampling-based \gls{mpc} and \gls{rl}, enabling control algorithms to search for optimal policies from an anthropomorphic prior.}
    \label{fig:teaser}
\end{figure}

Our contributions are as follows.
\begin{enumerate}
    \item We propose \gls{vqace}, a framework for embedding the action chunks of human hand motion into a quantized latent representation.
    \item We propose latent sampling ~\gls{mpc}, which is a real-time action synthesis algorithm that samples in the latent space.
    \item We propose action chunked \gls{rl}, which is a way to improve the exploration of \gls{rl} with action priors
\end{enumerate}

%% file: 02_related_works.tex
\section{Related Works}

\subsection{Dexterous Robot Manipulation Synthesis}

Traditionally, model-based approaches have demonstrated impressive results in controlling dexterous manipulation tasks across various platforms, such as quadrupeds~\cite{Yang2020WalkBall, Jean2023multicontactPlanning}, humanoids~\cite{Katherine2020BallJuggling}, drones~\cite{Dongjae2020DroneMPC}, and robotic hands\cite{lakshmipathy2024contactmpc}. 
Providing a powerful platform for hosting such complex robotic tasks, Tassa~et~al.~\cite{Howell2022mjpc} introduced MuJoCo MPC, a framework that simplifies the design and control of robotics problems through real-time predictive control.
However, these methods often involve a trade-off between using abstract models that simplify problem-solving and more accurate models that reduce modeling errors.

On the other hand, \gls{rl} bypasses the need for explicit dynamic models, learning policies directly from simulation. Although being successful in various tasks~\cite{Nikita2021rslgym, OpenAI2018Hand, Kaufmann2023RGPDrone}, it requires a significant amount of data and computational resources. Furthermore, transferring policies from simulation to real-world environments (sim-to-real) remains a persistent challenge.

The availability of large-scale datasets on robot grasping~\cite{Hidalgo2023AnthropomorphicGrasping, Yumeng2024RealDex, Hui2024GraspXl, Vuong2024Grasp-Anything, Lei2024MultiFingerGrasp, Ruicheng2023DexGraspNet} and manipulation tasks~\cite{Stephan2024dataOpenx, Khazatsky2024datadroid, Fang2024dataRH20T, Walke2023bridgedata} has facilitated progress in imitation learning, allowing impressive results in robot motion synthesis~\cite{Tony2023ACT, seungjae2024vqbet, Chi2023diffuionpolicy, Moo2024OpenVLA}.
These methods directly learn the mapping from observation to action in the dataset.

Our approach, \gls{vqace}, provides a middle ground between these methodologies. It is data-driven, leveraging neural network architectures akin to those used in imitation learning. However, instead of modeling the posterior distribution of actions conditioned on observations, it models the prior distribution of actions.  This prior distribution can provide guidance for the optimization of policy in \gls{mpc} and the exploration in \gls{rl}.

\subsection{Latents and Vector Quantization}

In the broader field of machine learning, latent representations are widely used to improve the quality of generation tasks~\cite{Rombach2021stablediffusion, Ramesh2022dalle}, and enhance semantic understanding~\cite{Radford2021Cliip}. Applied to robotics, Ghosh~et~al.~\cite{Octo2024} demonstrated cross-embodiment control by decoding latent vectors through embodiment-specific heads.  
Liconti~et~al.~\cite{liconti2024leveragingpretrainedlatentrepresentations} improved few-shot imitation learning by learning latent representations of actions.

Vector quantization of latents is another important technique, particularly in enhancing the performance of generative models across domains such as audio~\cite{prafulla2020jukebox}, images~\cite{Robin2021HighResimg}, video~\cite{Gai2024VQvideo}, and 3D reconstruction~\cite{navaneet2023compact3d}.
In robotic learning, vector quantization has also proven effective~\cite{Moo2024OpenVLA, Ankit2023RVT, seungjae2024vqbet, Muhammad2022Bet}.
Shafiullah~et~al.~\cite{shafiullah2022behavior} clustered continuous actions into discrete bins using k-means, a technique that enabled transformers to better predict discrete classes.

To address the scalability limitations of the k-means binning approach, 
Lee et al.~\cite{seungjae2024vqbet} employed residual vector quantization to their \gls{vae} network to encode actions into discretized vectors.
The method is successfully applied to imitation learning, achieving high performance and low computational cost compared to contemporary methods.
While their \gls{vae} network is unconditional, our \gls{vqace} is conditional—where actions are reconstructed based on the current joint positions. By conditioning on this additional context, our decoder achieves greater accuracy without increasing the amount of encoded information.

\begin{figure}[h]
    \centering
    \includegraphics[width=0.99\linewidth]{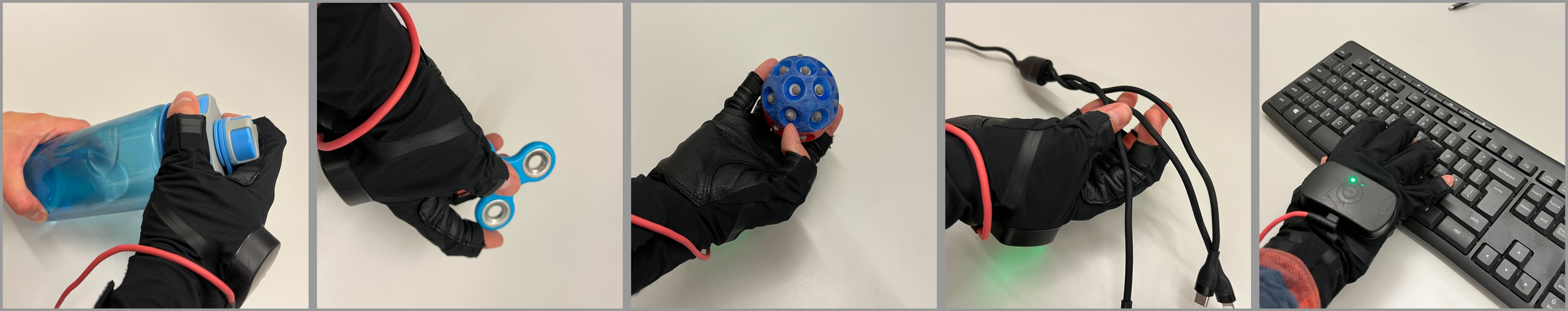}
    \caption{Samples of the tasks of data collection. The tasks covered everyday activities like bottle opening or keyboard typing.}
    \label{fig:data_collection}
\end{figure}

%% file: 03_methods.tex
\begin{figure*}
    \centering
    \includegraphics[width=\linewidth]{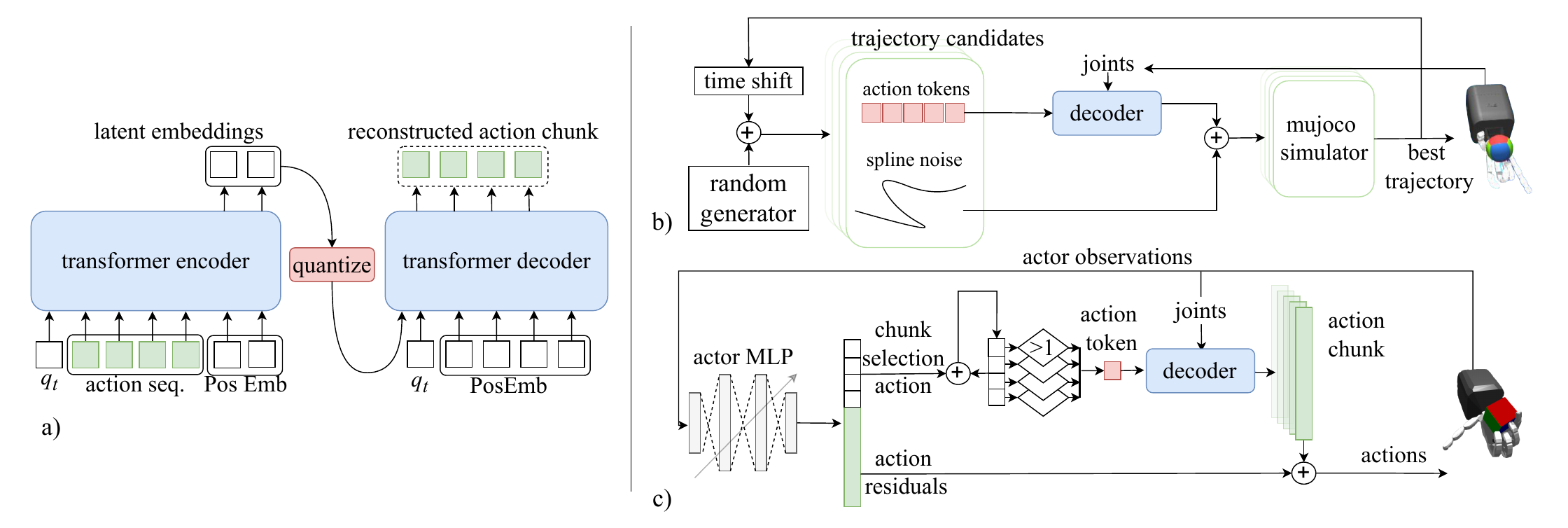}
    \caption{Architecture of \textit{Vector Quantized Action Chunking Embedding}(VQ-ACE) and its applications in sampling based \gls{mpc} and \gls{rl}. 
    a. \gls{vqace} (Sec.~\ref{subsec:vqace}). Conditioned on the current joint position, the encoder compresses the action sequence into a sequence of latent vectors. The latent vectors are quantized following the algorithm in \cite{Guyon2017vqvae}. 
    The decoder reconstructs the action sequence using both the joint position and quantized vectors. Causal masks are added to the decoder to allow time shifts in downstream applications.
    b. Latent sampling \gls{mpc} (Sec.~\ref{subsec:app_mjpc}). The control signal is the sum of the nominal policy decoded from latents and a Gaussian noise spline. The Mujoco simulator evaluates all sampled control sequences and selects the sequence with the best cost for execution. New control sequences are generated by applying time-shifted noise to the best sequence. Gaussian noise is added to the spline, and latent vectors are randomly flipped.
    c. Action chunked \gls{rl}  (Sec.~\ref{subsec:app_rl}). We augment the state and action spaces of a dexterous manipulation task with action chunk selection, which accumulates the agent's choices of action chunks and triggers the decoder. The decoder predicts the action chunk  for the next steps, using the current joint position and the latent vector selected by the actor. The final action output is the sum of the action chunk and residual values.}
    \label{fig:system}
\end{figure*}

\section{Action Chunking Embedding}
\label{sec:method}
In this section, we introduce \gls{vqace}, which embeds action chunks into latent spaces. We introduce the architecture and design decisions in Sec.~\ref{subsec:vqace}. 
After that, we present \textit{latent sampling \gls{mpc}} (Sec.~\ref{subsec:app_mjpc}) and \textit{action chunked \gls{rl}} (Sec.~\ref{subsec:app_rl}) where the latent space is utilized for searching or learning robot policies. an overview of the method can be found in Fig.~\ref{fig:system}.

\subsection{Data Collection}
\label{subsec:data_collection}

Our \gls{vqace} is trained with a human hand motion dataset that we collected with a motion capture glove~\cite{Rokoko}. We retargeted the hand pose into the 11 \gls{dof} of the robotic hand~\cite{Yasu2023GetBallRoll}, using the kinematic retargeting method from~\cite{sivakumar2022robotictelekinesislearningrobotic}. 
The dataset is 54 minutes long, recorded at 50 Hz, consisting hand motions of the interaction with daily objects and toys, cable lacing and sign languages. An example of the tasks acquired can be seen in Fig.~\ref{fig:data_collection}

\subsection{Vector-Quantized Action Chunk Embedding on behavior Data}
\label{subsec:vqace}

We employ a vector-quantized \gls{cvae} structure to learn a discrete latent embedding space for action chunks. 
As shown in Fig.~\ref{fig:system} a., we train a pair of encoder and decoder networks $\phi$, $\psi$.
The action chunk $a_{t:t+n}$, conditioned on the current joint position $q_{t}$, is encoded into a sequence of latent vectors $z_{k:k+m} = \phi(q_{t}, a_{t:t+n})$.
The latents are then quantized by the nearest-neighbor look-up in the codebook $z_q(z_k) = \operatorname{NearestNeighbor}_{\{e_1, e_2, e_3, ..., e_K\}}(z_k)$.
Conditioned on the current joint position $q_{t}$, the decoder reconstructs the action chunk ${\hat{a}}_{t:t+n}$ from $z_{k:k+m}$. 
Note that the latents and action chunks use two different sets of the indices $k$ and $t$, respectively. Both indices represent time, but they operate at different frequencies.
For simplicity, and with an abuse of notation, we define the mapping between the latent index and time as follows: 
\begin{equation} 
\label{eq:z2t} 
t(k) = \frac{k \cdot n}{m}
\end{equation}

Both the encoder and decoder are implemented using transformer architectures. 
The input of the \gls{cvae} encoder consist of tokens mapped from the current joint position $q_t$ and the action sequence $a_{t:t+n}$, prepended by a sequence of learned positional tokens. 
These learned tokens aggregate information from the entire input sequence, analogous to the [CLS] token in BERT~\cite{devlin2019bertpretrainingdeepbidirectional} but extended for sequence modeling.
The output features corresponding to these learned tokens are used to predicts the latents $z_{k:k+m}$, which serve as the input to the decoder.

The decoder, also based on a transformer architecture, takes as input the latents $z_{k:k+m}$, the current joint position $q_t$, and learned positional embeddings corresponding to the target output.
A causal mask is applied based on the time of each token rather than its position. 
The time of the latent variables $z_{k:k+m}$ is linearly mapped from their indices according to \eqref{eq:z2t}. This time-based masking ensures that the model respects temporal dependencies during the decoding process.

We use the following loss function $\mathcal{L}$ to train the networks
\begin{align}
\mathcal{L}_{\text {recon}} & =\left\|a_{t: t+n} - \psi\left(q_{t}, z_q\left(\phi\left(q_{t},a_{t: t+n}\right)\right)\right)\right\|_1 \\
\mathcal{L}_{\text {commit}} & =\left\|\phi\left(q_{t},a_{t: t+n}\right)\right) -  \mathrm{SG}\left[z_q\left(\phi\left(q_{t},a_{t: t+n}\right)\right)\right] \|_2^2 \\
\mathcal{L} & =\mathcal{L}_{\text {recon }} + \lambda_{\text {commit }}  \mathcal{L}_{\text {commit}},
\end{align}
where $\mathrm{SG}$ is the stop gradient operator. 
For vector quantization, we used the \gls{ema} version of the approach presented in~\cite{Guyon2017vqvae}, which updates the embedding vectors using \gls{ema} rather than an auxiliary loss function. This modification has been shown to be capable of stabilizing training in practice~\cite{sonnet_vqvae}. 

\subsection{Predictive Sampling in Latent Space}
\label{subsec:app_mjpc}

Predictive sampling is a straightforward \gls{mpc} algorithm that iteratively improves its action sequences by sampling $N$ candidate variants of current plan, simulating them, and updating the plan with the best candidate.
Thanks to its simplicity and intuitive nature, it serves as an ideal testing ground for validating and comparing the quality of representations from which action sequences are sampled.
We hypothesize that the better the structure of the latent representations, the higher the likelihood of the algorithm discovering good policies and achieving superior performance. 

A high-performance implementation of the algorithm is provided in~\cite{Howell2022mjpc} .
Using the original algorithm as a baseline, where action candidates are represented by splines in each control dimension, we validate the structure of the latent space by modify the algorithm to sampling from the latent space according to \eqref{eq:randomlatents} and \eqref{eq:randomspline}.

For each task, we specify a cost function $J$ and aim to minimize this cost. Each plan $\Pi$ is represented by latents $z_{k:k+m}$, and a noise spline $\theta_{\tau:\tau+P}$, where we use the shorthand $\theta=\theta_{\tau:\tau+P}$. Similar to \eqref{eq:z2t}, the indicies of the spline points are also aligned in time with $t(\tau) =  \frac{\tau \cdot n}{P}$. 
Given a query time $\tilde{t}$ and the current joint positions $q_t$, the control signal $u$ from the policy is evaluated as follows:
\begin{equation}
    u(\tilde{t}) = \psi\left(\tilde{t}; q_t, z_{k:k+m}\right) + s\left(\tilde{t}; \theta_{\tau:\tau+P}\right),
\end{equation}
where the first term represents the action at $\tilde{t}$ reconstructed from the decoder, and $s$ is the spline evaluation as described in~\cite{Howell2022mjpc}.

Predictive sampling improves policies by generating multiple samples around the current policy and selecting the best, avoiding gradient-based optimization. In latent space, candidates are modified by replacing elements with probability $p$, selecting new codes from the codebook, enabling local search and progressive optimization.
This process is formalized as follows:
\begin{equation}
\label{eq:randomlatents}
z_j^{(i)}= \begin{cases}
z_j, & \text { with probability } 1-p \\ 
e_r, & \text { with probability } p, \\ 
     & \text { \; where } r \sim \operatorname{Uniform}(1, K)
\end{cases},
\end{equation}
where $j\in \{k, k+1, \dots, k+m\}$,  $z_j$ is the current value of the latent variable, $z^{(i)}_j$ is the $i'th$ sample, and $e_r$ is the new value sampled uniformly from the codebook.

In addition to the actions , we also perturb the noise spline to offer finer-grained control over the sequence. The noise spline is reset to zero whenever the latents are modified.
\begin{equation}
\label{eq:randomspline}
\theta^{(i)} \sim \begin{cases}
\mathcal{N}\left(\theta, \sigma^2\right), & \text { if } z^{(i)} = z\\ 
\mathcal{N}\left(0, \sigma^2\right), & \text { if } z^{(i)} \neq z
\end{cases}.
\end{equation}

In our experiments, the latents represent only the actions of the robotic hand. For tasks that involve control over other \gls{dof}s (such as a robotic arm), we apply latent sampling exclusively to the hand's \gls{dof}, while using the original predictive sampling algorithm for the other \gls{dof}s.

\subsection{RL with Action Chunks}
\label{subsec:app_rl}

\gls{rl} is a powerful technique built upon the concepts of states, their dynamics, and observations. However, the use of action chunks subtly contradicts the Markovian assumption of traditional \gls{rl}.
The action priors modeled by our \gls{vqace}, typically spanning 10 to 20 timesteps per latent token, are too long to be considered as actions of a single state, yet too short to serve as subgoals in hierarchical \gls{rl}~\cite{Pateria2022HRL}. 
In this section, we present \textit{action chunked \gls{rl}}, a simple method that enables \gls{rl} to learn to utilize action chunks and facilitate policy exploration. 

We let the action chunk $A_t$ be the nominal actions and let the agent adjust it with a residual $\delta_t$. 
For a system with dynamics $x(t+1) = f(x(t), u(t))$, we augment the states $x$, action $u$, and the dynamics $f$ as:
\begin{equation}
    \hat{x}(t+1) = \hat{f}\left(\hat{x}(t), \hat{u}(t)\right),
\end{equation}

with
\begin{align}
    \hat{x}(t) &= [x(t); A_t; x_s(t)], \\
    \hat{u}(t) &= [\delta_t; u_s(t)].
\end{align}
The action chunk $A_t$ is not updated at every time steps, so we define it as $A_t :=  a_{t_a:t_a+n}$, $a_t \leq t$. And it is used as the input of the original system,
\begin{equation}
    u(t) = \begin{cases}
        a_t + \delta_t, & \text{ if } t_a \leq t < t_a+n \\
        a_{t_a+n} + \delta_t & \text{otherwise}
    \end{cases}.
\end{equation}

We call $x_s \in \mathbb{R}^K$ and $u_s \in \mathbb{R}^K$  the chunk selection states and the chunk selection actions respectively. As the input of the chunk decoder is quantized with a codebook of size $K$, these vectors represent the agent's selection of the action chunk. $a_t$, the feedforward action to be executed at time $t$, and $x_s(t)$ are included in the actor's observation for it to make decisions. The dynamics of chunk selection is as follows:
\begin{align}
\operatorname{trigger}_t  &=  \mathbf{1}_{\{ \max\left(x_s(t) + u_s(t)\right) > 1 \}}\\
A_{t+1} &= \begin{cases}
    \psi\left(q_t, \operatorname{argmax} (x_s(t) + u_s(t)) \right) & \text{ if } \operatorname{trigger}_t \\
    A_t & \text{ otherwise } 
\end{cases} \\
x_s(t+1) &= \begin{cases}
     \operatorname{Uniform}(0, 1)  & \text{ if } \operatorname{trigger}_t \\
    x_s(t)+u_s(t) & \text{ otherwise } 
\end{cases} 
\end{align}

The chunk selection states accumulate the the chunk selection actions for each iteration. When any of $x_s$ is greater than 1, it triggers the update of the action chunk. The new action chunks are used as nominals in the following steps, and the $x_s$ is reset. 
This design bridges the frequency difference between the multi-step action chunk execution and the single-step residual feedback. Moreover, it enables the policy to actively select the action chunk to execute.

%% file: 04_experiments.tex
\begin{figure*}
    \centering
    \includegraphics[width=0.9\linewidth]{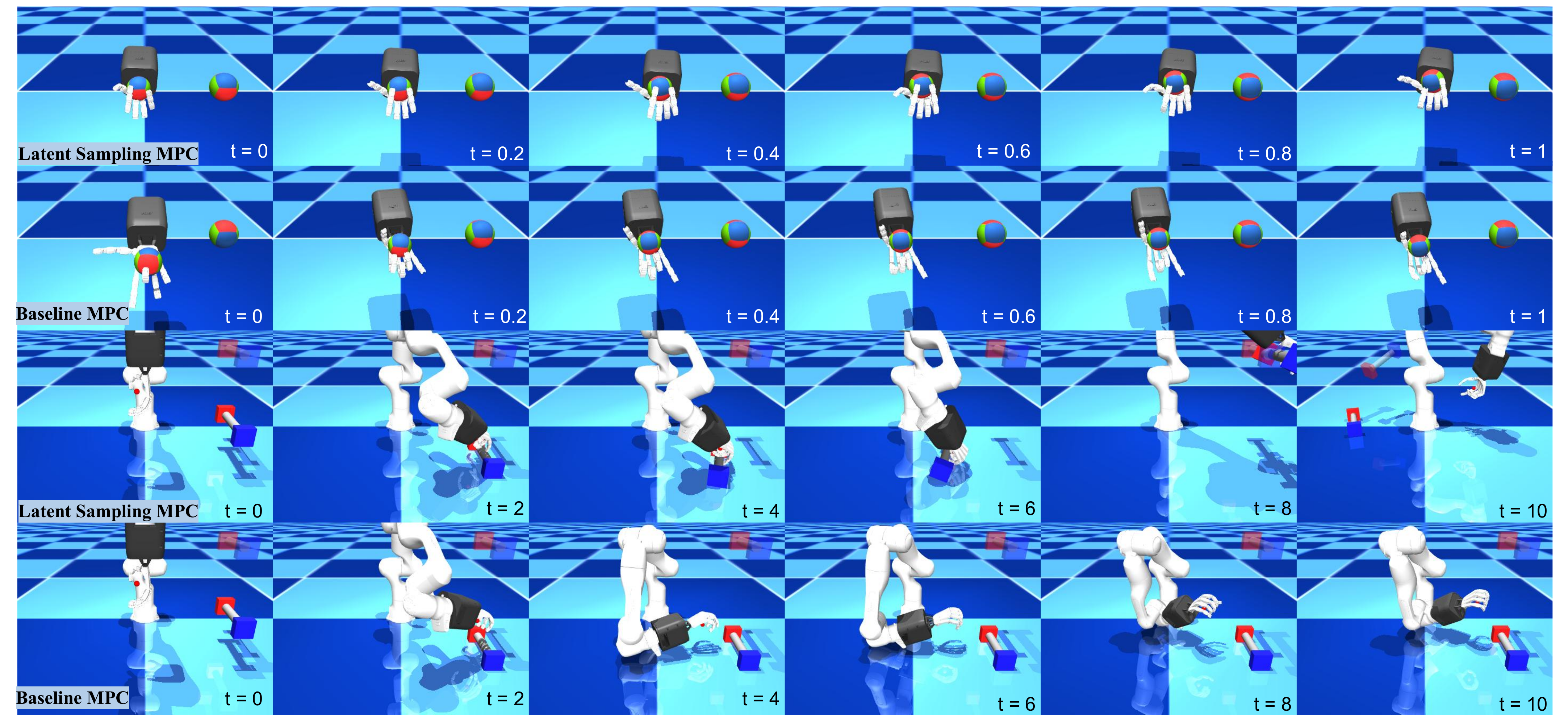}
    \caption{The first two rows show the execution of the Ball Rolling task. The screenshots are taken from with a time step of 0.2 seconds. In each figure, the left side shows the robotic hand in simulation, and the right side shows the orientation of the reference ball. 
    We observe that our latent sampling \gls{mpc} generates more human-like behavior, with all fingers maintaining contact with the ball. 
    In contrast, the baseline sampling-based MPC produces more arbitrary actions, as it merely selects the best control that drives the ball toward the target pose.
    The two bottom rows represent the Object Picking task. These snapshots are taken at intervals from 0s to 10s with a timestep of 2s. The transparent object represents the target pose. 
    When the object is carried to a position within a threshold of the target, the task is considered successful, and both the object and the target pose are randomized for the next trial. 
    In the third row (latent sampling \gls{mpc}), the robot successfully grasps the object at 6 seconds and lifts it at 8 seconds. At 10s, the object and target positions are updated due to the robot's success. On the other hand, the baseline sampling-based MPC fails to grasp the object and gets stuck around the target without a successful attempt (last row).}
    \label{fig:mujoco_visulization}
\end{figure*}
\section{Experiments}

\subsection{Action Space Embedding}

With our \gls{vqace}, we encode an action chunk of 1 second (50 timesteps) into 5 tokens. Each latent token has a dimension of 16 and is quantized using a codebook with a size of 4. 
Both the encoder and decoder networks consist of 3-layer transformers, each with 4 attention heads, a model dimension of 128, and a feedforward dimension of 512. The network is trained with a commitment loss weight of 0.1, a learning rate of 1e-4, and a batch size of 3072. The L1 loss of the trained model on validation dataset converges to 0.050.

\subsection{Latent Sampling MPC}

We validate our \gls{vqace} and the learned latent space by sampling based \gls{mpc}.
The algorithm is described in Sec.~\ref{subsec:app_mjpc}
We validate our \gls{vqace} and the learned latent space through sampling-based MPC, focusing on two tasks: \textbf{Ball Rolling} (in-hand manipulation) and \textbf{Object Picking} (grasping and placing a dumbbell-shaped object). 
For the Ball Rolling task, the controller manages the robotic hand's pose and joint positions to rotate a ball along the x-axis while keeping it near a target position. The hand can make slight adjustments using gravity. In the Object Picking task, the robot must pick up a dumbbell-shaped object from a random position and place it in a random target location and orientation.

These tasks test the dexterity of a \textit{Faive} biomimetic  robotic hand presented in ~\cite{Yasu2023GetBallRoll} . The hand contains a total of 16 rolling contact joints, of which 11 are actuated \gls{dof}s. A 7-\gls{dof} Franka Emika 3 arm is used for the picking task. Sampling controllers were run on an Intel i9-14900K CPU, while neural network inferences were handled by an NVIDIA 4090 GPU. In our setup, the simulation and planning processes are executed synchronously. We ensure that the simulation process runs at the same speed for both our method and the baseline ~\cite{Howell2022mjpc}. Note that this approach accounts for computational overhead, penalizing planners with higher computational costs, as they lead to a lower control frequency during simulation and potentially worse performance.
The horizon of both planners are set to 1.0s and both these algorithms are tuned separately for performance.
A series of screenshots are provided in Fig.~\ref{fig:mujoco_visulization}. 

\begin{figure}
    \centering
    \includegraphics[width=0.9\linewidth]{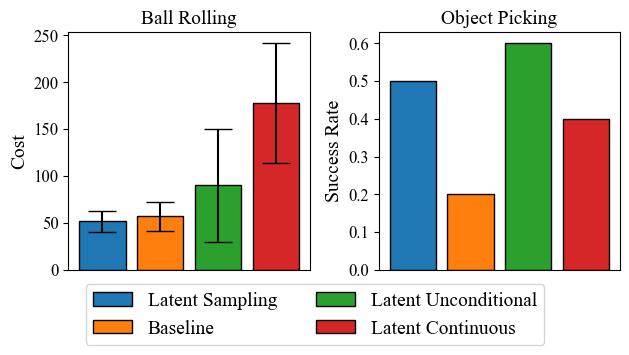}
    \caption{Cost and success rate in the Ball Rolling and Object Picking tasks. We compare our proposed latent sampling with the baseline predictive sampling and two ablations. The values in the histogram represent the mean, and the error bars represent the standard deviation of ten runs}
    \label{fig:ablation}
\end{figure}

We compare the two algorithms quantitatively in terms of cost (lower is better) and task success rate (higher is better). We test our method and the baseline in both tasks sampling 40 trajectories, and the results are summarized in Fig.~\ref{fig:ablation}, from which we can see that our \gls{vqace} out performs the baseline in both tasks.

\begin{figure}
    \centering
    \includegraphics[width=0.9\linewidth]{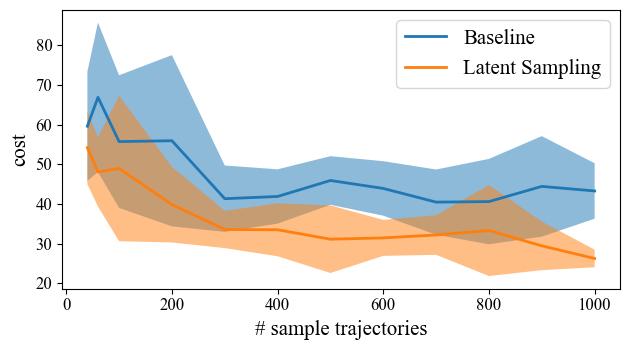}
    \caption{The cost vs number of sampled trajectories. The solid line is the mean while the area shade represents the standard deviation. Each of the configuration is repeated for 7 time. The cost of latent sampling \gls{mpc} is always less than the baseline given the same number of trajectories sampled, showcasing the benefit of our method. Moreover, when the number of trajectories is 40 our latent sampling \gls{mpc} achieves the cost (54.2) that the baseline as at 200 trajectories (56.0), suggesting that our latent sampling mpc can drastically reduce the number of samples needed to achieve the same level of performance}
    \label{fig:n_traj_vs_cost}
\end{figure}

Since the sampling-based \gls{mpc} selects control actions from randomly generated candidates, a larger number of sampled trajectories increases the likelihood of finding optimal results.
Therefore, we tested the algorithms adjusting the number of trajectories sampled per iteration. 
As the computational cost rises with more sampled trajectories, we adjusted the simulation speed accordingly, ensuring that the controller is updated at a frequency between 30Hz and 50Hz. As shown in Fig.~\ref{fig:n_traj_vs_cost}, the cost has a decreasing trend for both the baseline sampling \gls{mpc} and the latent sampling \gls{mpc}. 
However, the cost of latent sampling \gls{mpc} is always less than the baseline given the same number of trajectories sampled, suggesting that our latent sampling \gls{mpc} samples from a better space of action parameterization.

\subsection{Action Chunked RL}

We applied Action Chunked \gls{rl} to the following tasks:

\textbf{Cube Reorientation}: This task tests in-hand manipulation dexterity. The objective is to reorient a 50mm cube in-hand to a target pose without dropping it. The performance metric used is consecutive successful reorientations.

\textbf{Cube Stacking}: In this environment, the \textit{Faive} hand is mounted on a 7-\gls{dof} Franka Emika 3 robotic arm. The task is considered successful if a small cube with a 50mm side length is stacked on top of a larger cube with a 70mm side length.

For both tasks, we defined straightforward cost functions and trained the policies using PPO~\cite{Schulman2017PPO}. The cost functions remained consistent between our method and the baseline to ensure fair comparisons. Training was parallelized across 4096 environments and executed on an NVIDIA 4090 GPU.

\begin{figure}
    \centering
    \includegraphics[width=0.9\linewidth]{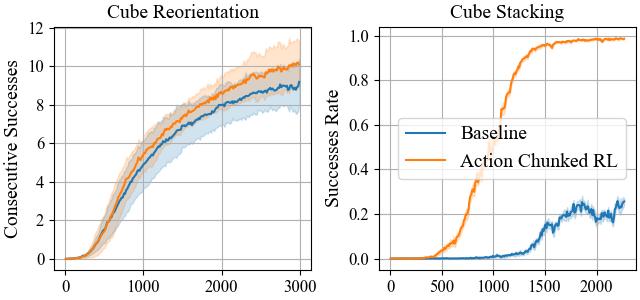}
    \caption{Performance comparison of Action Chunked RL and Baseline on Cube Reorientation (Left) and Cube Stacking (Right). Means (solid lines) and standard deviations (shaded areas) are computed from 5 runs with different random seeds. Action Chunked RL demonstrates faster convergence and better final performance in both tasks.}
    \label{fig:rl_curves}

\end{figure}

Fig.~\ref{fig:rl_curves} illustrates the consecutive successes and success rate for the cube reorientation and cube stacking tasks, respectively. The results demonstrate that our Action Chunked \gls{rl} approach leads to faster convergence and superior performance in the cube reorientation task. Furthermore, it significantly improves performance in the cube stacking task, where the baseline struggles to find an effective policy.

This improvement can be attributed to the fact that Action Chunked \gls{rl} enables the agent to explore meaningful action priors, as opposed to the random action exploration typical in baseline methods.

\subsection{Ablation Studies}

\gls{vqace} employs action chunking and vector quantization to model the natural motion of human hands. Conditioned on the current joint position, it translates discrete latent code into actions in the future. In this section, we discuss how do these components contribute the effectiveness of \gls{vqace}. 

\paragraph{Conditional observation}

Our model learns the distribution of future action chunks conditioned on the current joint positions $q_t$. 
This conditioning provides additional information that facilitates the reconstruction of action chunks. Instead of capturing only the absolute hand posture, the latent codes are able to capture the relative movement from the current states.
Ablating $q_t$ from the input of the encoder and decoder increases the validation L1 error from 0.05 to 0.07.
As shown in Fig.~\ref{fig:ablation}, the cost of ball rolling with unconditional variants is much higher than our latent sampling approach. 
However, the unconditional decoder performs better on the object-picking task, as it tends to focus on encoding static hand postures rather than dexterous in-hand manipulation.

\paragraph{Vector quantization}

We use vector quantization to model multimodal action distributions. Without vector quantization, a \gls{vae} can also be used to reduce the action's search space. As an ablation study, we trained models using KL loss instead of vector quantization.  
Conditioned on $q_t$, The decoder predicts the action chunk from 6-dimensional continuous vectors, which represent half of the controllable \gls{dof}s. 
We apply the decoder in a sampling-based \gls{mpc}, where nominal and sampled trajectories are represented and explored as splines in latent spaces. We tuned the exploration noise for performance while keeping the other configurations consistent with both our method and the baseline.

The \gls{vae} reconstructs the action chunk more accurately, achieving an L1 loss of 0.028 compared to 0.050 of the quantized variants.
However, as shown in Fig.~\ref{fig:ablation}, it performs worse than quantized versions in both tasks, particularly in the ball-rolling task.

Based on the results, we hypothesize that the VAE ablation may be more prone to overfitting compared to vector-quantized alternatives.

%% file: 05_conclusion.tex
\section{Conclusion}
In this work, we presented \gls{vqace}, a novel approach to addressing the complexity and high dimensionality of dexterous robotic manipulation tasks. By embedding sequences of human hand motion into a quantized latent space, we significantly reduced the action space's dimensionality while preserving essential motion characteristics.
Our approach was validated through integration with both sampling \gls{mpc} and \gls{rl}, showing more effective exploration and policy learning. 
In \gls{mpc} experiments, we proposed latent sampling, which produced more human-like behaviors, yielding higher task success rates and lower costs in tasks like Ball Rolling and Object Picking. 
Similarly, in \gls{rl} experiments, our action chunked \gls{rl} accelerated convergence and improved exploration in cube stacking and in-hand cube reorientation tasks.

\subsection{Limitations}
One limitation of \gls{vqace} is that action chunks are represented as sequences of nominal joint positions, which restricts the decoder to a specific embodiment. Additionally, the method focuses on feed-forward actions, relying on feedback from a downstream controller. Another limitation lies in the relatively small size of the in-house dataset, though this can be mitigated by leveraging larger public datasets with accurate hand pose estimations.

\subsection{Future Work and Possible Applications}
The success of \gls{vqace} in simulation and real-world settings underscores its potential for broader applications. Beyond robotic manipulation, this framework offers promising implications for managing large state spaces in other fields, such as legged locomotion, or humanoid robots. By providing a scalable solution for handling complex, high-dimensional tasks, \gls{vqace} contributes to advancing the capabilities of adaptive and generalist robotic systems. 